# Using Taint Analysis and Reinforcement Learning (TARL) to Repair Autonomous Robot Software

D. Lyons and S. Zahra
Robotics & Computer Vision Laboratory
Dept. of Computer & Information Science
Fordham University
New York 10458 USA

*Abstract*—It is important to be able to establish formal performance bounds for autonomous systems. However, formal verification techniques require a model of the environment in which the system operates; a challenge for autonomous systems, especially those expected to operate over longer timescales. This paper describes work in progress to automate the monitor and repair of ROS-based autonomous robot software written for an a-priori partially known and possibly incorrect environment model. A taint analysis method is used to automatically extract the data-flow sequence from input topic to publish topic, and instrument that code. A unique reinforcement learning approximation of MDP utility is calculated, an empirical and non-invasive characterization of the inherent objectives of the software designers. By comparing off-line (a-priori) utility with on-line (deployed system) utility, we show, using a small but real ROS example, that it's possible to monitor a performance criterion and relate violations of the criterion to parts of the software. The software is then patched using automated software repair techniques and evaluated against the original off-line utility.

## I. INTRODUCTION

As robot systems are increasingly deployed in domestic, industrial and military applications, it is increasingly important to develop a method to establish bounds for how systems will perform under realistic operating conditions. While there is an established engineering tradition of using mathematics to model physical structures and processes sufficiently to guarantee that a construction will fulfill its specification, there are challenges applying this to autonomous systems [1]. Not many software developers have the skill set necessary to formally establish program correctness according to a specification. But even if they did, formal verification of software [2] [3] and autonomous systems [4], as well as 'correct by construction' formal synthesis methods [5] *all* suffer from the problem that they require a designer-specified model of the environment. And that model can be incomplete because of designer oversight or oversimplification of a complex and dynamic environment. For autonomous systems that are expected to operate over longer timescales, there is also the issue of encountering significant drift over time between the model and the environment encountered by the agent.

Inspired by progress in empirical automated software repair [6] we propose an approach to the automatic analysis and repair of autonomous robot system software to handle environment model related performance failures. Leveraging static analysis techniques [7] used in compilers and security analysis, the approach does not require software engineers to do more than they are doing now in terms of coding. It aims to accomplish for autonomous systems, what approaches like SapFix [8], Clearview [9] and others [6] have achieved in automated software patching.

The remainder of the paper is laid out as follows: Section II introduces the basic ideas in our approach. Section III uses a short example to illustrate our non-invasive approach to analysis and repair. Section IV summarizes the direction of the work and the issues of scaling to future plans.

## II. TARL APPROACH

The first assumption of our proposed approach is that while most autonomous robot software is not developed using formal verification techniques, nonetheless, software engineering experience is such that the systems produced by experienced software developers are a solid first approximation to a system with formal performance guarantees. We will begin our approach to formal performance guarantees by leveraging the skills of software developers. This has the advantages of not asking the software development community to adopt new formalisms, and widening the scope of our method, at least in principal, to include any existing (ROS) robot software.

Our second assumption is that experienced software professionals are capable of writing a performance monitor that will signal when their software is working as they intended it to work (i.e., to their apprehension of the specification). The field of runtime verification addresses this capability formally [10], but our need is for far less; an informal monitor would suffice.

It is extremely common for the developers of autonomous robot software to test their algorithms using a simulation tool [11]. While is it true that simulation tools do not replace physical testing, especially for uncertain environments, they nonetheless have become very powerful and comprehensive – driven often by the needs of realistic gaming applications. It is very likely still true that "Simulation is doomed to succeed" [12] in the sense that simulations only model what the designer expects to happen (or v.v.). However, software designers' use of simulation opens for us an approach to non-invasive and automatic formal characterization of working software, for at least the conditions that the simulation implements. And we will show that ultimately this will not be a restriction.

### A. Formal Model

An autonomous robot system operating in a partially known environment will be modeled as an MDP $M(S, A, P, R)$ where the state $S$ is a function of sensor signals (we will augment this presently), the actions $A$ are the actuator commands, the transition function $P$ is the stochastic environment, and reward $R$ is the feedback on whether the system is satisfying its performance specification. A policy $\pi$ for $M$ selects which action to carry out based on the state. We will consider the human-written system software to implicitly embody the policy for $M$, and the human-written performance monitor to implicitly embody the reward R.

The utility $U(s)$ $s \in S$ for an MDP is a measure of the long-term desirability of each $s \in S$ and can be written via a Bellman eq:

$$U(s_i) = R(s_i) + \gamma \sum_j P(s_j | s_i, \pi(s_i)) U(s_j) \quad (1)$$

for discount $\gamma$ and deterministic policy $\pi$. We propose that utility presents a non-invasive and automatic way to characterize the performance of autonomous system software, to indicate where the software may be lacking when the performance monitor fails, and to automatically repair the software. Reinforcement learning [13] gives us a way to calculate utility, but only if the software can be automatically instrumented for the calculation.

### B. ROS

The ROS robot operating system is extremely widely used and has greatly improved the portability of robot code since its release in 2007. A ROS system is composed of computing *nodes* that communicate over named channels called *topics*. Sensor driver software *publishes* incoming measurements on sensor topics. Other computing nodes can *register* callback functions that will be triggered whenever data arrives on a topic. Actuator drivers, that control the mechanism, register callbacks that are triggered whenever data is written to actuator topics. We will restrict our attention to ROS software both because of its popularity and its convenient topic/node structure. A ROS computation graph shows computation nodes connected via the topics they communicate on. This makes it very clear where data flows are coming in and going out.

### C. Taint Analysis and Reinforcement Learning (TARL)

The performance monitor R, written by the human software designers, tests all the input topics necessary to determine if the performance guarantee holds. It is necessary to find an automatic way to identify the connections, line by line through computing node source code, between input topics and actuator topic publishing commands. While some of this data flow is evident in the ROS computation graph, most of it is hidden *within the source code*. We will automatically find these data flows and instrument them so that the utility can be estimated in terms of sensor state and statement number. We will then show how this information can be used to detect when software is no longer performing correctly due to unanticipated aspects of the environment, and to potentially repair it.

Taint analysis [7] is a static analysis technique employed in compiler optimization and security analysis. The analysis starts with a program and storage location to investigate for a potential attack, the *sink*. Any user input location is a taint *source*. The program is investigated to see what additional storage locations become tainted in storing any computations that involve a tainted location and whether this leads to the sink. The sequence of statements between source and sink is the *taint list*. We use PyT [14] (modified for ROS) to find taint lists in ROS/Python code and instrument them for utility learning.

```
20 def callback(data): # collect position sensor
21     global pos
22     pos = data.pose.pose.position # from topic
23
24 def travel(goal,vout): # move to goal
25     global pos
26     err, delta, vel = 1, 0, 0
27     while err>Epsilon: # stop within Epsilon
28         delta = goal – pos
29         err   = abs(delta)
30         vel   = 5 * delta
31         vout.publish(vel) # write to topic
32
33 if __name__ == '__main__':
34   try:
35     rospy.init_node('Traveller', anonymous=True)
36     rospy.Subscriber(Odometry,callback)
37     vpub=rospy.Publisher(Velocity,Twist,10)
38     while True: # go back and forth
39        travel(G1,vpub)
40        travel(G2,vpub)
41   except rospy.ROSInterruptException:
42     pass
```

Taint List:
('if __name__=='__main__'', 33)
('data=Odometry', 36)
('pos=data.pose.pose.position',22)
('while True', 38)
('delta=goal – pos',27)
('vel=5*delta',30)
('vout.publish(vel)',31)

Figure 1: ROS/Python program implementing a recurrent robot task (top); output of ROS-modified PyT taint analysis (bottom).

We use this as follows: Each taint list line is automatically instrumented with a hook function that calls a reinforcement learning (RL) algorithm. To calculate the utility (called *value* in RL) we ask the software designers to repeatedly run the software, *once they consider it finished*, using whatever simulation they have been using. Whatever tradeoffs the designers implicitly made in implementing the specification is made explicit in this utility and will be our tool for understanding the software and potentially repairing it. We employ a SARSA TD-RL [13] algorithm to estimate utility in terms of sensor state, taint-list line number *and* a reward-scaled flow value. This last unique addition to the SARSA rule makes utility sensitive not just to control-flow, but also to *how* the information is being manipulated at each line.

## III. EXAMPLE

Figure 1(top) shows python code for a ROS node that shuttles a robot between two goal locations (e.g. delivery robot). The performance monitor (not shown) is that the robot reaches each goal within a time and precision bound. The taint list (Fig. 1(bottom)) tags the data flow in the program and is instrumented for RL estimation of utility as described. For convenience, the data below shows just the G1 to G2 transit.

### A. Offline Utility

Let us consider that the robot has an odometry sensor and velocity output, and also a binary terrain type sensor. This latter is unused in Fig.1 as the designers consider the terrain won't affect the performance. Fig. 2 shows the utility calculated by the SARSA algorithm converged for the performance specification (reward) for the 'off-line' case, i.e., before the software is deployed. The red points are the terrain sensor true case, and blue are the false case. Line number are the left axis and odometry the right. This graph is an empirical characterization of the software designers' intent, subject to perhaps faulty information about the environment.

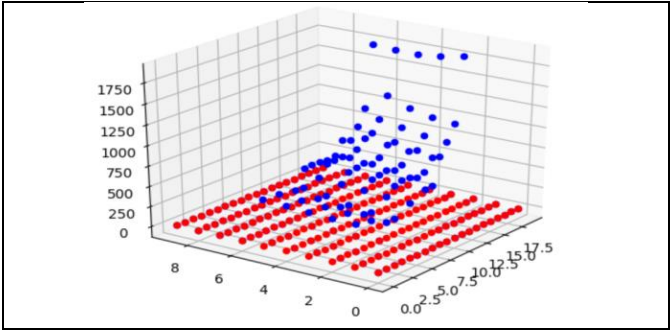
Figure 2: Utility table run under expected environment conditions

### B. On-line Utility

Now let us consider what happens when our autonomous robot system software is deployed, and it is found that the environment does not behave as expected. It is found that muddy terrain conditions are sometimes experienced in transit, which slows down the progress of the robot towards its G2 goal. Using the same RL framework, utility from real life missions is collected; some missions experience the terrain issues and (sometimes) fail and some experience the same environment the designers expected. In this paper, to simplify the time aspects of testing, we don't run a physical robot mission. We run another version of the off-line simulation in which the environment model is changed.

Fig. 3 shows the utility collected during missions which sometimes failed because of unexpected and unmodeled terrain conditions. The horizontal axis is odometry and the colors are as before. Looking at the difference of off-line and on-line utilities we can isolate a region of *maximum utility difference*, indicated in Fig 3 by the dashed rectangle. In that region we calculate the average utility curves for the taint list on-line and off-line cases and compare them

$$qs_\pi(n) = \sum_{m,p \,\in\, diff} q_\pi(m,p,n) \quad (2)$$

Where *n* is line number, *m* is terrain sensor, *p* is odometry and $q_\pi$ is the utility.

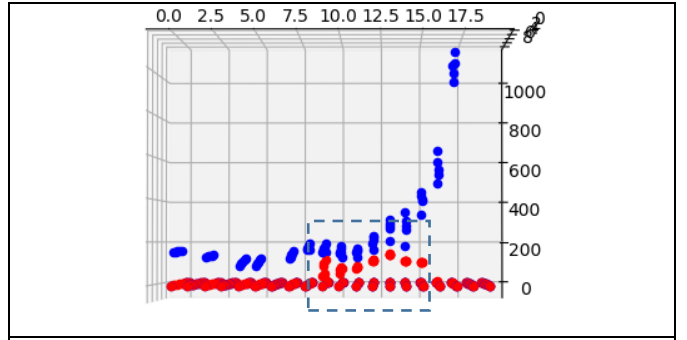
Figure 3: Utility table under unexpected environment conditions

Figure 4(a) shows this comparison, which evidences a peak at the 2$^{nd}$ last line (vel=5*delta). As it shows the maximum change in utility, we propose this identifies it as the most likely culprit for the failure and, hence, the candidate for repair analysis.

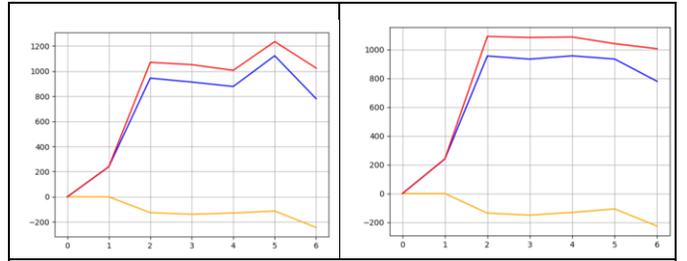
Figure 4: Comparison of off-line (blue) and on-line (yellow) utility slices $qs_\pi$ with difference (red) (a) original, (b) lines swapped.

It is reasonable to ask if the peak simply precedes the line closest to the reward (the publish). To show that our addition of flow information to the SARSA rule addresses this, the same test was done but with Fig.1 lines 29 and 30 swapped. The result is shown in Fig. 4(b) which now has a peak at the 3$^{rd}$ last line, correctly reflecting the swap. The addition of the reward-scaled flow value to the SARSA update equation is what makes utility sensitive to the values being calculated.

### C. Automated Repair

Automated program repair [6] is an area of increasing interest to automate and speed up the patching of commercial software bugs, especially security related bugs and for *self-healing* autonomous devices. The process is typically divided into two phases: determining where the bug might be located in the software and proposing/testing fixes to the bug. Our approach here will address the single line culprit model we developed in the prior section. However, there is nothing that would limit this being applied to a multiple line culprit model. We will also continue to use RL as a solution method – this time to search

over the space of program modifications. Determining how to edit the code will be divided into two steps:
1. Mutation of the affected code to generate potential software patches
2. A modified ε-Greedy policy and SARSA TD-RL algorithm to determine which mutation produces optimal performance

We use the Python AST library to determine any *constants* in the identified line and generate a set of mutated values for these constants [8]. Every copy of the mutated instruction is *protected* (i.e., within an IF statement) by the condition used to make the value table slices, Eq. (2). Rather than testing one mutation at a time, all mutations are tested by selecting among them with an ε-Greedy policy and monitoring average total reward (*ATR*) compared to that of the initial, off-line performance

$$ATR = \sum_t R_t / E \qquad (3)$$

where $E$ is the number of episodes and $\sum_t R_t$ the total reward up to time $t$. Fig. 5 shows a graph of *ATR* for the off-line case (blue), the on-line case that encounters problems (yellow), the ε-Greedy search of mutations for repair (green), and the final selected repair (red). The *ATR* performance of the ε-Greedy search, while it is better than the on-line performance, is not as good as the original off-line performance. The reason is that the ε-Greedy modification conducts significant search exploration of the mutated code, picking up a lot of performance hits. The curve is slowly rising and looks as if it will approach the offline value eventually. However, once we estimate the optimal mutation, we can simply re-estimate the utility using only that optimal mutation. That is the function shown in red on the graph, re-establishing the performance guarantee.

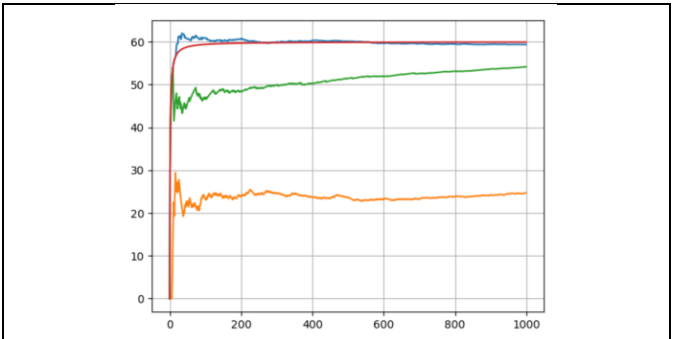

Figure 5: Code mutation and evaluation for the running example

## IV. CONCLUSIONS

While it is becoming evident that its necessary to establish performance guarantees for autonomous systems, this is very challenging in practice. Software designers rarely have the mathematical skills to use significant formal verification. And even if they did, uncertainty about the environment encountered in execution (their specification for verification) would weaken this approach this in any case. We are inspired by the extremely practical work done in automated software repair [8] to propose an automated analysis and repair approach for existing software for autonomous robot systems.

This paper describes work in progress, and the results presented are from a thorough but small feasibility study. Our next steps are to apply this to a large quadrotor mission in ROS-M. Issues of scale need to be addressed in a mission with a multipart performance guarantee resulting in multiple taint lists of varying lengths being identified. By calculating one utility per taint list and using the full multi-part performance monitor as reward for every list, we propose to avoid the potential exponential increase in utility size ($2^{num\_lists}$) while allowing for performance related cross-effects between taint flows. The example here uses a tabular utility. We expect however to need topic-specific, parametric forms in future work.

## V. REFERENCES

[1] R. Simmons, C. Pecheur and G. Srinivasan, "Towards Automatic Verification of Autonomous Systems," in *Intelligent Robots and Systems (IROS) 2000*, Takamatsu Japan, 2000.

[2] C. Baeir and J.-P. Katoen, Introduction to Model Checking, Cambridge MA: MIT Press, 2008.

[3] R. Jhala and R. Majumdar, "Software Model Checking," *ACM Computing Surveys,* vol. 41, no. 4, 2009.

[4] D. Lyons, R. Arkin, S. Jiang, M. O'Brien, F. Tang and P. Tang, "Performance Verification for Robot Missions in Uncertain Environments," *Rob. & Aut. Systems,* vol. 98, pp. 89-104, 2017.

[5] H. Kress-Gazit and G. Pappas, "Automatic Synthesis of Robot Controllers for Tasks with Locative Prepositions.," in *IEEE Int. Conf. on Robotics and Automation*, Anchorage, Alaska, 2010.

[6] C. LeGoues, M. Pradel and A. Roychoudhury, "Automated program Repair," *CACM,* vol. 12, pp. 56-65, December 2019.

[7] D. Boxler and K. Walcott, "Static Taint Analysis Tools to Detect Information Flows," in *16th Int'l Conf. Software Eng. Research and Practice (SERP'18)*, Las Vegas, NV, 2018.

[8] A. Marginean, J. Bader, S. Chandra, M. Harman, Jia Y, M. K, M. A. and A. Scott, "SapFix: Automated End-to-End Repair at Scale," in *41st Int. Conf. on Soft. Eng.*, Montreal Canada, 2019.

[9] J. Perkins, S. Kim, S. Larsen and e. al, "Automatically Patching Errors in Deployed Software," in *SIGOPS 22nd Symp. on Operating Systems Principles*, 2009.

[10] E. Bartocci, Y. Falcone, A. Francalanza and G. Reger, "Introduction to Runtime Verification.," in *Lectures on Runtime Verification LNCS Volume 10457*, Springer Cham, 2018.

[11] T. Erez, Y. Tassa and E. Todorov, "Simulation Tools for Model-Based Robotics," in *IEEE Int. Conf. Rob. & Aut*, Seattle WA, 2015.

[12] R. Brooks and M. Mataric, "Real Robots, Real Learning Problems," in *Robot Learning*, Springer US , 1993.

[13] R. Sutton and A. Barto, Reinforcement Learning, Cambridge MA: MIT Press, 2018.

[14] B. Thalman, "PyT: A Static Analysis Tool for Detecting Security Vulnerabilities in Python Web Applications," Aalborg Univ Denmark, 2016.